\pdfoutput=1
\documentclass[10pt, a4paper]{article}
\usepackage{lrec2022} 
\usepackage{multibib}
\newcites{languageresource}{Language Resources}
\usepackage{graphicx}
\usepackage{tabularx}
\usepackage{soul}
\usepackage{titlesec}
\titleformat{\section}{\normalfont\large\bfseries\center}{\thesection.}{1em}{}
\titleformat{\subsection}{\normalfont\SmallTitleFont\bfseries\raggedright}{\thesubsection.}{1em}{}
\titleformat{\subsubsection}{\normalfont\normalsize\bfseries\raggedright}{\thesubsubsection.}{1em}{}
\renewcommand\thesection{\arabic{section}}
\renewcommand\thesubsection{\thesection.\arabic{subsection}}
\renewcommand\thesubsubsection{\thesubsection.\arabic{subsubsection}}

\usepackage{epstopdf}
\usepackage[utf8]{inputenc}

\usepackage{dirtytalk}
\usepackage{hyperref}
\usepackage{xstring}
\usepackage{adjustbox}
\usepackage{booktabs}
\usepackage{arydshln}
\usepackage{multirow}
\usepackage{amsfonts}
\usepackage{amsmath}
\usepackage{float}
\usepackage{placeins}

\usepackage{todonotes}
\usepackage{color}

\usepackage{xcolor}

\makeatletter
\newcommand\footnoteref[1]{\protected@xdef\@thefnmark{\ref{#1}}\@footnotemark}
\makeatother

\renewcommand{\vec}[1]{\ensuremath{\mathbf{#1}}}

\DeclareMathOperator*{\argmax}{argmax~}

\title{Czech Dataset for Cross-lingual Subjectivity Classification}

\name{Pavel P\v{r}ib\'{a}\v{n}$^{1,2}$, Josef Steinberger$^{1,2}$} 

\address{University of West Bohemia, Faculty of Applied Sciences, \\
        $^{1}$NTIS -- New Technologies for the Information Society,\\
        $^{2}$Department of Computer Science and Engineering, \\
        Univerzitn\'{i} 2732/8, 301 00 Pilsen, Czech Republic \\
         \{pribanp,jstein\}@kiv.zcu.cz\\
         \tt \url{https://nlp.kiv.zcu.cz}\\}

\abstract{In this paper, we introduce a new Czech subjectivity dataset of 10k manually annotated subjective and objective sentences from movie reviews and descriptions. Our prime motivation is to provide a reliable dataset that can be used with the existing English dataset as a benchmark to test the ability of pre-trained multilingual models to transfer knowledge between Czech and English and vice versa. Two annotators annotated the dataset reaching 0.83 of the Cohen’s $\kappa$ inter-annotator agreement. To the best of our knowledge, this is the first subjectivity dataset for the Czech language. We also created an additional dataset that consists of 200k automatically labeled sentences. Both datasets are freely available for research purposes. Furthermore, we fine-tune five pre-trained BERT-like models to set a monolingual baseline for the new dataset and we achieve 93.56\% of accuracy. We fine-tune models on the existing English dataset for which we obtained results that are on par with the current state-of-the-art results. Finally, we perform zero-shot cross-lingual subjectivity classification between Czech and English to verify the usability of our dataset as the cross-lingual benchmark. We compare and discuss the cross-lingual and monolingual results and the ability of multilingual models to transfer knowledge between languages.
\\ \newline }

\begin{document}

\maketitleabstract




\section{Introduction}
Subjectivity classification \cite{wiebe-etal-1999-development} is one of the integral parts of sentiment analysis (opinion mining). Its basic purpose is to determine if a sentence or phrase is subjective or objective \cite{liu2012sentiment}. It can be further used to improve other tasks such as polarity detection or information extraction \cite{wiebe-etal-1999-development,english-dataset}. Nowadays, the subjectivity classification is often used as a benchmark test \cite{zhao2015self-adasent,reimers-gurevych-2019-sentence,wang2021entailment-few-shot-learner,bragg2021flex} in transfer learning to test abilities and language understanding of pre-trained BERT-like language models based on the Transformer architecture \cite{attention-all-transformer}.

\par The goal of subjectivity classification is to classify a sentence or a clause of the sentence as subjective or objective. Subjective sentences express personal feelings, views, beliefs or opinions and objective sentences hold or describe some factual information \cite{liu2012sentiment}.

\par Evaluation of the pre-trained models for transfer learning is a crucial part of their development. For English, the well-known GLUE \cite{wang-etal-2018-glue} and SuperGLUE \cite{superglue} benchmarks are available. These benchmarks contain a set of diverse tasks that allow a thorough evaluation of English pre-trained models. 

\par For multilingual models such as mBERT \cite{devlin-etal-2019-bert} or XLM-R \cite{xlm-r}, the XTREME \cite{pmlr-v119-hu20b-xtreme} benchmark can be used to test their ability of cross-lingual transfer learning and knowledge transfer between languages. Unfortunately, the XTREME benchmark does not include any task for the Czech language.


\par Our main motivation is to partly fill this gap and contribute a bit by introducing a reliable Czech dataset that can be used for the cross-lingual evaluation. We intend to use the dataset to test the cross-lingual abilities of pre-trained multilingual models in pair with the existing English dataset \cite{english-dataset} as a benchmark for zero-shot cross-lingual subjectivity classification. Thus, partly test the ability of pre-trained multilingual models to transfer knowledge between Czech and English. We are aware that to properly evaluate any pre-trained model, a diverse set of tasks is needed, but we believe that even one task can be helpful in the evaluation process. To the best of our knowledge, there is no subjectivity dataset for the Czech language, therefore our secondary goal is to extend the available dataset resources for Czech.

In this paper, we present the first Czech dataset for subjectivity classification task that consists of 10k manually annotated sentences from movie reviews and movie descriptions. Secondly, we provide an additional dataset of 200k sentences labeled in a distant supervised way (automatically). The automatic labeling is based on the idea from \cite{english-dataset} that movie reviews contain mostly subjective sentences and the movie descriptions usually consist of objective sentences. We describe the process of building and annotating the dataset. The dataset is annotated by two annotators and the Cohen’s $\kappa$ \cite{cohen1960coefficient} inter-annotator agreement between them reaches $0.83$. We perform experiments with two multilingual mBERT \cite{devlin-etal-2019-bert} and XLM-R-Large \cite{xlm-r} and three monolingual Transformer based models on the new Czech dataset and providing a competitive baseline of $93.56\%$ of accuracy. Next, we conduct experiments with the same two multilingual models on the English dataset to be able to compare our cross-lingual experiments. Our results for the monolingual experiments with English are on par with the current state-of-the-art results. Finally, we evaluate the multilingual models and their ability to transfer knowledge between English and Czech on the zero-shot cross-lingual classification task. The cross-lingual experiments show that using only English data for fine-tuning the XLM-R-Large, the model can achieve worse results only by $2.8\%$ on the Czech dataset compared to the model trained on Czech data. When the model is trained using only the Czech data, the result on the English dataset is roughly $4.4\%$ worse than the current state-of-the-art results.


Our main contributions are the following: 1) we introduce the first Czech subjectivity dataset that allows cross-lingual evaluation in pair with the existing English dataset. 2) We perform a series of monolingual and cross-lingual experiments. We set a competitive baseline for the new Czech dataset. We compare abilities of two multilingual models to transfer knowledge between Czech and English in the subjectivity classification task. 3) We release\footnote{\label{foot:github}The datasets and code are freely available for research purposes at \url{https://github.com/pauli31/czech-subjectivity-dataset}} the dataset and code freely for research purposes, including the dataset splits for easier comparison and reproducibility of our results.

\section{Related Work}
\label{sec:related-work}

The subjectivity classification task was a popular research topic at the beginning of the 21st century. It was studied in many papers \cite{wiebe-wilson-2002-learning,wiebe2004learning,exploiting-2005-subjectivity,esuli-sebastiani-2006-determining,wiebe-mihalcea-2006-word,mihalcea-etal-2007-learning}. Nowadays, the subjectivity classification is often used as a benchmark for the evaluation of pre-trained models intended for transfer learning.

In \cite{wiebe-etal-1999-development}, the authors describe the annotation process of 1k news sentences. Four annotators annotated the sentences as subjective or objective, but because some sentences can be considered borderline examples, they also assigned certainty ratings, ranging from 0, for least certain, to 3, for most certain.  We use special label trash for the borderline sentences during our annotation, see Section \ref{subsubsec:annotation-procedure}.

\par \newcite{english-dataset} created English subjectivity dataset from movie reviews and movie descriptions. They automatically made the dataset using the assumption that reviews have mostly subjective sentences and descriptions usually contain objective sentences. The resulted dataset consists of 10k sentences, see Table \ref{tab:data-stats}. Further in this paper, we reference the dataset as the English dataset.

\par In terms of Czech resources, the Czech subjectivity lexicon \textit{Czech SubLex 1.0} \cite{sublex} contains a list of words with assigned sentiment polarity and part-of-speech tags. 

\par There are also pairs of existing datasets that can be used for the cross-lingual evaluation similarly to our approach. For example, the Czech sentiment dataset of movie reviews \textit{CSFD} \cite{habernal-etal-2013-sentiment} can be used with the English \textit{IMDB} \cite{maas-etal-2011-learning-imdb} sentiment reviews dataset as shown in \cite{priban-steinberger-2021-multilingual}. Another example is the multilingual corpus \cite{piskorski-etal-2019-second} for named entity recognition (NER) that contains labels in the same format for four Slavic languages, including Czech. Next, the Czech aspect-based sentiment dataset \cite{hercig2016unsupervised-absa} can be evaluated together with the English dataset \cite{pontiki-etal-2014-semeval} and both of them come from the same domain and contain the same set of labels.

\par The initial work focused on cross-lingual subjectivity classification is presented in \cite{mihalcea-etal-2007-learning}. The authors investigated methods to automatically generate resources for subjectivity analysis for new language by using a parallel corpus and available resources in English. \newcite{amini-etal-2019-cross} performed cross-lingual subjectivity classification between English and Persian. Other work that is related to cross-lingual subjectivity can be found in \cite{saralegi2013cross}.

\par In \cite{wang2021entailment-few-shot-learner}, the authors used the English subjectivity dataset as one of the tasks to evaluate their approach for few-shot learning based on RoBERTa model \cite{liu2019roberta}. \cite{nandi2021empirical-indove-subj} analyzed various models for text representation, including the original BERT model \cite{devlin-etal-2019-bert} on the English dataset. Similarly, in \cite{zhao2015self-adasent,CNN-MCFA-subj-2018,khodak-etal-2018-la-byte-mLSTM7,reimers-gurevych-2019-sentence}, the authors also used the English dataset to evaluate the performance of their newly designed models.

\section{Dataset Building}
We provide two datasets
of subjective and objective Czech sentences from movie reviews and movie descriptions (plot summaries), respectively. We use the mentioned idea from \newcite{english-dataset}, in which the authors automatically created English dataset (\texttt{Subj-EN}) of 10k subjective and objective sentences. They assume that the descriptions are mostly objective and the reviews are subjective. This assumption is valid in most cases, but there can also be objective sentences in reviews and subjective sentences in descriptions. The number of these noisy samples differs in both cases, as you can see in Table \ref{tab:annotations-stats}.

\par For this reason, we decided to create a manually annotated dataset (\texttt{Subj-CS}) of 10k examples that should eliminate the incorrect occurrences as much as possible.
Secondly, we automatically built an additional dataset (\texttt{Subj-CS-L}) of 200k sentences using almost the same approach\footnote{Based on our observations in the dataset, we decided to use sentences or phrases with at least six tokens but they used sentences longer than nine tokens.} as in \cite{english-dataset}.

\subsection{Cleaning and Obtaining Data}
We acquired roughly 4M reviews and 735k descriptions from Czech Movie Database\footnote{\url{https://www.csfd.cz}} (CSFD) during October 2021. The Czech sentiment movie review dataset \cite{habernal-etal-2013-sentiment} also consists of reviews from CSFD. We assume that in the future, our dataset can be used in combination with the sentiment dataset therefore, we decided to remove the sentiment reviews from the data downloaded by us. We were able to match and remove about 74k reviews out of a total of 91k from the sentiment dataset. The remaining 17k reviews were most likely changed or removed from the CSFD website since the authors of the sentiment dataset originally downloaded the data in 2013.

\par\textcolor{black}{Next, we split the reviews and descriptions into sentences by UDPipe 2} \cite{straka-2018-udpipe}\footnote{We use the \textit{czech-pdt-ud-2.5-191206.udpipe}  model.}. We have to note that in some cases, it failed to split the sentences correctly, especially for sentences without a space after the first sentence.

\begin{figure}%
    \centering
\includegraphics[width=8.1cm]{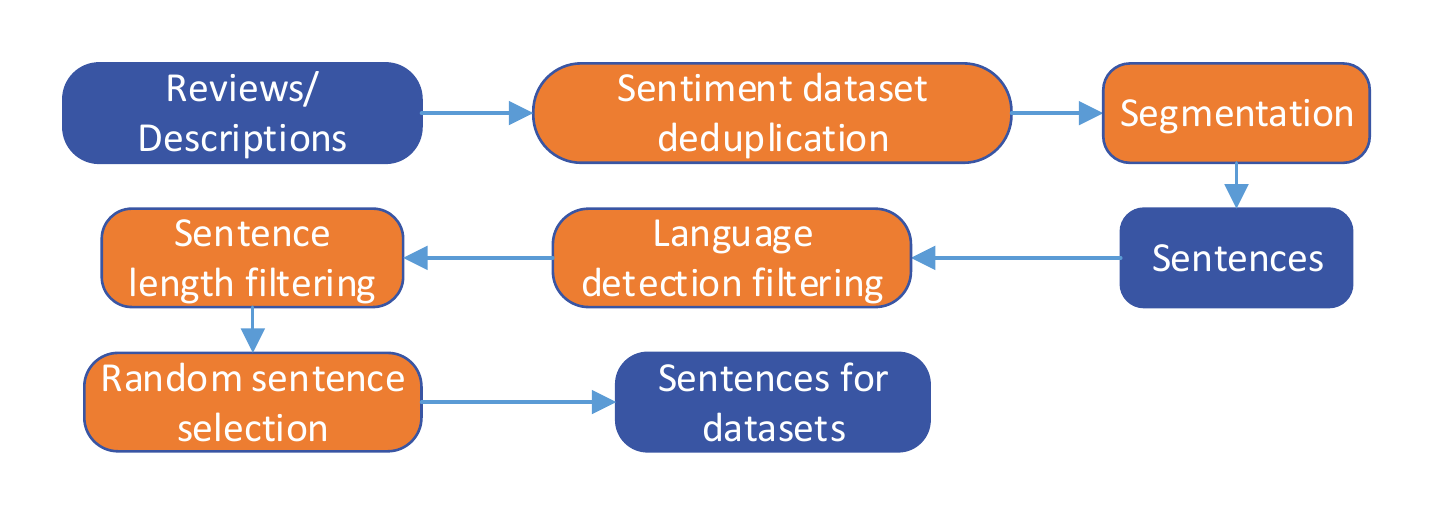}
\caption{Data cleaning pipeline visualization.} 
\label{fig:cleaining-pipeline}
\end{figure}

The reviews can contain phrases instead of grammatically correct sentences, but we do not distinguish between them. Some of the texts (mostly reviews) were written in other languages (most often Slovak and English). We filter these out\footnote{We use the Python package \texttt{langdetect} available at \url{https://pypi.org/project/langdetect/} to detect the language.} and we keep only Czech sentences. Finally, we filter out sentences with less than six tokens. See Figure \ref{fig:cleaining-pipeline} for the cleaning pipeline visualization.

\par The entire cleaning process resulted in 884k and 19M sentences (phrases) from descriptions and reviews, respectively. We randomly selected 40k sentences from the obtained reviews and descriptions for manual annotation and 200k sentences (100k from reviews and 100k from descriptions) for the automatically created dataset. The remaining sentences are not used.

\subsection{Data Annotation}
In this section, we describe the process of manual annotation. Two Czech native speakers performed the annotation. Even though the subjectivity classification may seem like an easy task, it showed to be rather difficult for some sentences to assign a subjectivity label.

\subsubsection{Annotation Procedure}
\label{subsubsec:annotation-procedure}
Firstly, the task of subjectivity classification was explained to the annotators along with the meaning of the subjective and objective sentences according to the definition in \cite{liu2012sentiment}. The annotators were also asked to read the papers mentioned in Section \ref{sec:related-work} to clearly understand the task. We summarize the annotation guidelines in Section \ref{subsec:annotation-guideline}.

\par \textcolor{black}{During the first annotation stage,} each of the annotators was asked \textcolor{black}{to label a common set of} 100 sentences with one of three labels: \texttt{subjective, objective} and \texttt{trash}, see Section \ref{subsec:annotation-guideline} for their description. 

\par We use the trash label because, despite our best data cleaning efforts, there were still undesirable texts: e.g., short sequences of words that does not make any sense (random words), only numbers and other characters, sentences in other languages, texts that were obviously incorrectly segmented and made no sense etc.

After the first 100 annotated sentences, the annotators discussed the conflicts to clarify and improve the annotation guidelines. Based on the discussion, we decided to extend the annotation labels by two more \texttt{unsure} and \texttt{question}.

\par \textcolor{black}{The questions appeared to be rather problematic. The subjectivity was not clear very often and thus, we decided to exclude them.} In addition, the questions are only in a tiny part of the data, i.e., $1.73\%$ and $2.41\%$ for review and description sentences, respectively, see Table \ref{tab:annotations-stats}.



\par The unsure label was added because for some sentences, the annotators were not able to assign the subjectivity. For example,
sentences for which a context (previous sentence) is needed to decide, sentences that describe a movie or event, but contain some clearly subjective adjective(s) and they can be perceived or interpreted both as subjective or objective depending on an individual person. Other problematic sentences are commands, wishes or parts of poems and rhymes. Here we list some of the problematic sentences that both annotators labeled with the unsure label:

\begin{itemize}
	\small
	\item[] {(1) \say{\textit{Všechno ovšem tak snadné řešení nemá.}} -- \say{\textit{Not everything has such an easy solution.}}}
	
	\item[] {(2) \say{\textit{To je dobrý důvod pro to, aby byla Japonsku vyhlášena válka.}} -- \say{\textit{That's a good reason to declare war on Japan.}}}
	
	\item[] {(3) \say{\textit{Dnes večer je to však díky napjaté atmosféře velmi obtížné.}} -- \say{\textit{Tonight, the tense atmosphere makes it very difficult.}}}
	
	\item[] {(4) \say{\textit{Drastický horor, při kterém tuhne krev v žilách}} -- \say{\textit{Drastic horror that makes your blood run cold}}}
	
	\item[] {(5) \say{\textit{Tak se o to postará příroda sama!}} -- \say{\textit{Nature will take care of it!}}}
\end{itemize}

\par We decided to add these additional labels because we wanted to assign labels only in cases where the annotators are very confident with their annotations and thus obtain more reliable annotations without controversial examples and dataset of high quality.   

\par After the update of the annotation guideline, both of the annotators assigned labels to the same 2,034 sentences. The Cohen’s $\kappa$ \cite{cohen1960coefficient} inter-annotator agreement for this 2k sentences reaches $0.68$ for all five labels. Because we provide the dataset only with the objective and subjective labels, we exclude any sentence with at least one\footnote{Each sentence has two labels -- one from each annotator.} of the trash, unsure or question labels. Thanks to this filtration, we obtain 1,668 sentences only with the subjective and objective labels. The Cohen’s $\kappa$ for this subset is $0.83$, which represents a fairly good level of agreement. The remaining 141 conflict sentences are then resolved with the help of third person.

\par Finally, almost 5,000 sentences were annotated by each of the two annotators resulting in a total of 11,907 annotated sentences, see Table \ref{tab:annotations-stats}. We can see that the subjective and objective sentences are relatively balanced in the annotated samples and we believe that this reflects the real data distribution. Even though we obtained more than 5,000 sentences with the subjective and objective labels, we cut the annotations to have exactly 5,000 examples for each of the two labels. We decided to provide a perfectly balanced dataset since it allows easier comparison and evaluation of experiments. In our experiments, we use only the sentences with the subjective and objective labels, i.e., 10,000 sentences. We refer to this dataset as \texttt{Subj-CS}.

The entire procedure of annotation can be summarized into the following steps:

\begin{enumerate}
    \item Each annotator annotated 100 sentences as \texttt{subjective}, \texttt{objective} or \texttt{trash}.
    
    \item Every conflict in the first 100 sentences was discussed separately between the annotators to clarify and improve the annotation guideline. We extended the annotation guideline by two more labels: \texttt{unsure} and \texttt{question}.
    
    \item 2,034 sentences are annotated by each annotator (1,668 as subjective or objective with 141 conflicts). The Cohen’s $\kappa$ reaches $0.83$ for subjective and objective sentences. The conflicts are resolved by a third person.
    
    \item Almost 10k another sentences are annotated in total by both annotators. The annotations are cut down to contain exactly 5,000 subjective and objective sentences.
\end{enumerate}

\subsubsection{Annotation Statistics}
The manual annotation resulted in a total of 11,907 annotated sentences with one of five labels, see Table \ref{tab:annotations-stats}.

\par \textcolor{black}{During the annotation procedure, we set the limit of at most 15 review sentences for the same movie and at most three description sentences in the 40k sentences selected for the manual annotation. However, }the average number of sentences for the same movie is only $1.43$ and $1.02$ for review and description sentences, respectively.

\begin{table}[h!]
\catcode`\-=12
\begin{adjustbox}{width=0.90\linewidth,center}
\begin{tabular}{lrrr}
\toprule
Label      & \multicolumn{1}{c}{Reviews} & \multicolumn{1}{c}{Descriptions} & \multicolumn{1}{c}{Total}   \\ \midrule
unsure     & 866 /\  \small{13.11\%}               & 457 /\ \ \   \small{8.62\%}                     & 1 323                       \\
\textbf{object.} & \textbf{726 / \small{10.99\%}}               & \textbf{4 464 / \small{84.22\%}}                  & \textbf{5 190}                        \\
\textbf{subj.} & \textbf{4 794 /\  \small{72.57\%}}              & \textbf{208 /\ \ \   \small{3.92\%}}                     & \textbf{5 002}                        \\
quest.   & 114 /\ \ \ \small{1.73\%}              & 128 /\ \ \   \small{2.41\%}                     & 242                         \\
trash      & 106 /\ \ \  \small{1.60\%}               & 44 /\ \ \   \small{0.83\%}                      & 150                        \\ \cdashline{1-4}
Total      &  6 606 / \ \ \  \small{100\%}    & 5 301 / \ \ \  \small{100\%}                    & 11 907                      \\
\bottomrule   
\end{tabular}
\end{adjustbox}
\caption{Annotation statistics for subjective and objective} \label{tab:annotations-stats}
\end{table}

\par As we assumed, a considerable percentage of sentences in reviews are not subjective (only $72.57\%$ of sentences are subjective). Similarly, there is also a relatively large part of sentences in the movie descriptions that are not objective ($84.22\%$ of the sentences are objective).


\subsubsection{Annotation Guideline}
\label{subsec:annotation-guideline}
The annotators were instructed to annotate a given sentence or phrase with one of five labels. Based on the subjectivity description from \cite{wiebe-etal-1999-development,english-dataset,liu2012sentiment}, the sentence should be annotated as subjective if it expresses or evokes some personal feelings, views, beliefs or the sentence holds an opinion about entities, events or their properties (mostly movies in our case) from the non-objective point of view. For example:
\begin{itemize}
	\small
	\item[] { \say{\textit{Samotný film se mi líbil, ale nepřekvapil.}} -- \say{\textit{I liked the movie itself, but it didn't surprise me.}}}
\end{itemize}

\par The sentence should be annotated as objective if it contains some factual information about an entity, event or their properties but does not hold a personal or subjective opinion about it and it does not try to convince or impose some opinion to the reader, for example:
\begin{itemize}
	\small
	\item[] { \say{\textit{Maurice žije a pracuje v jižní Francii.}} -- \say{\textit{Maurice lives and works in the south of France.}}}
\end{itemize}

\par The disputed and controversial sentences, sentences where the annotator is not sure about its subjectivity or sentences for which context from previous text is needed to decide should be annotated with the unsure label, see Section \ref{subsubsec:annotation-procedure} for examples. The trash label is used for sentences or phrases that do not make any sense or contain random words, characters or numbers. The question label is used for sentences that are questions.

\subsection{Automatic Dataset}
Besides the manually annotated dataset, we also built a large dataset (named \texttt{Subj-CS-L}) in a distant supervised way using the same approach as in \cite{english-dataset}. We labeled 100k review sentences as subjective and 100k movie description sentences as objective ones. All sentences have to have at least six tokens. We believe that even if the dataset contains some incorrect labels, it could be useful in combination with the manually created dataset, for example, in an unsupervised pre-training.


\section{Data \& Models for Experiments}
\label{sec:data-models}
For the experiments, we split the \texttt{Subj-CS} dataset into three parts with the following ratio: $75 \%$ for training, $5 \%$ for the development evaluation and $20 \%$ for testing.
For the cross-lingual experiment with the \texttt{Subj-CS-L} dataset from Czech to English, we use $5 \%$ as the development evaluation data and the rest is used for training.

\par Because there is no official split for the English dataset \cite{english-dataset}, we use 10-fold cross-validation for the monolingual experiments to be able to compare our results with other papers. We also split the English dataset into training, development and testing parts with the same test size (see Table \ref{tab:data-stats}) that was used in \cite{wang2021entailment-few-shot-learner}\footnote{Unfortunately, they do not provide any script or details to obtain the identical split. In other words, we do not know which sentences belong to the training part and which to the testing part.}. For all three Czech and English datasets, we provide a script to obtain exactly the same data split to allow reproducibility and future comparison of our results.

\begin{table}[ht!]
\catcode`\-=12
\begin{adjustbox}{width=0.95\linewidth,center}
\begin{tabular}{lrrrr}
\toprule
Dataset                 & \multicolumn{1}{l}{Name}  & \multicolumn{1}{c}{Subjective}         & \multicolumn{1}{c}{Objective}                               & \multicolumn{1}{c}{Total} \\ \midrule
\multirow{4}{*}{Subj-CS}   & \multicolumn{1}{l}{cs-train} & 3 750   & 3 750                 & 7 500                    \\
                        & \multicolumn{1}{l}{cs-dev}   & 250  & 250                 & 500                      \\
                        & \multicolumn{1}{l}{cs-test}  & 1 000    & 1 000                 & 2 000                     \\ \cdashline{2-5}
                        & \multicolumn{1}{l}{} & 5 000   & 5 000                & 10 000                     \\ \midrule
\multirow{3}{*}{Subj-CS-L}     & \multicolumn{1}{l}{cs-L-train} & 95 000    & 95 000                  & 190 000                      \\
                        & \multicolumn{1}{l}{cs-L-dev}   & 5 000    & 5 000                     & 10 000                       \\ \cdashline{2-5}   
                        &\multicolumn{1}{l}{} & 100 000    & 100 000                     & 200 000                    \\ \midrule
\multirow{4}{*}{Subj-EN} & \multicolumn{1}{l}{en-train} & 3 764  & 3 736                  & 7 500                    \\
                        & \multicolumn{1}{l}{en-dev}   & 231   & 269                      & 500                     \\
                        & \multicolumn{1}{l}{en-test}  & 1 005   & 995                     & 2 000                     \\ \cdashline{2-5}
                        & \multicolumn{1}{l}{} & 5 000  & 5 000                & 10 000                    \\  \bottomrule            
\end{tabular}
\end{adjustbox}
\caption{Datasets statistics.}
\label{tab:data-stats}
\end{table}

\subsection{Transformer Models}
For the experiments, we use solely the pre-trained BERT-like models based on the encoder part of the original Transformer architecture \cite{attention-all-transformer}. The modified language modeling task is used to pre-train all the models, see the corresponding papers for details.

\par We employ three Czech monolingual models \textit{Czert-B} \cite{czert}, \textit{RobeCzech} \cite{straka2021robeczech}, \textit{Czech Electra} model \cite{czech-electra}, two multilingual models \textit{mBERT} \cite{devlin-etal-2019-bert}, \textit{XLM-R} \cite{xlm-r} and the original monolingual English \textit{BERT} model \cite{devlin-etal-2019-bert}, see Table \ref{tab:models-size} for their size (in a number of parameters) comparison.

\begin{table}[ht!]
\catcode`\-=12
\begin{adjustbox}{width=0.75\linewidth,center}
\begin{tabular}{lrrr} 
\toprule
Model        &  \multicolumn{1}{c}{\#Params} & \multicolumn{1}{c}{Vocab} & \multicolumn{1}{c}{\#Langs} \\ \midrule
Czech Electra & 13M        & 30k   & 1                             \\
Czert-B      & 110M       & 30k   & 1                             \\
RobeCzech    & 125M    & 52k   & 1                                 \\
BERT   & 110M       & 30k  & 1                             \\
mBERT        & 177M       & 120k  & 104                           \\
XLM-R-Large  & 559M       & 250k  & 100                          \\
\bottomrule
\end{tabular}
\end{adjustbox}
\caption{A comparison of used models: number of parameters, vocabulary size and a number of supported languages.}
\label{tab:models-size}
\end{table}


\paragraph{Czech Electra} \cite{czech-electra} is Czech model based on the Electra-small model \cite{clark2020electra}.

\paragraph{Czert-B} \cite{czert} is Czech variant of  the original BERT\textsubscript{BASE} model \cite{devlin-etal-2019-bert}.

\paragraph{RobeCzech} \cite{straka2021robeczech} is Czech version of the RoBERTa model \cite{liu2019roberta}.

\paragraph{BERT} \cite{devlin-etal-2019-bert} is the original BERT\textsubscript{BASE} model.

\paragraph{mBERT} \cite{devlin-etal-2019-bert} is a cased multilingual version of the BERT\textsubscript{BASE} that was jointly trained on 104 languages. 

\paragraph{XLM-R-Large} \cite{xlm-r} is a multilingual version of the RoBERTa \cite{liu2019roberta} that supports 100 languages.

We fine-tune all the models for the binary classification task, i.e., subjective vs. objective sentence detection. For all models based on the original BERT model, we use the hidden vector $\vec{h} \in \mathbb{R}^H$ of the classification token \texttt{[CLS]} that represents the entire input sequence, where $H$ is the hidden size of the model. The vector is obtained from the pooling layer, i.e., from a fully-connected layer of size $H$ with a hyperbolic tangent used as the activation function. The dropout of $0.1$ is  applied and the result is then passed into a task-specific linear layer represented by matrix $\vec{W} \in \mathbb{R}^{|2| \times H}$. The output class $c$ (subjective or objective) is computed as
$c = \argmax(\vec{h}\vec{W}^{T})$.

\par For the XLM-R-Large and RobeCzech models, the same\footnote{The first artificial token \texttt{<s>} of the input sequence is used instead of the \texttt{[CLS]} token.} approach is used and in addition, an extra dropout of $0.1$ is applied before the pooling layer (as in the original RoBERTa implementation).
\par We use the Adam \cite{Kingma-adam} optimizer with default parameters ($\beta_1 = 0.9, \beta_2 = 0.999$) and the cross-entropy loss function.

\section{Experiments}
To set baseline results for the new Czech dataset and verify its usability as a cross-lingual benchmark dataset between Czech and English, we performed a series of experiments with Transformer based models. The experiments can be categorized into two groups -- \textit{monolingual} and \textit{cross-lingual}.

\par In monolingual experiments for Czech, we fine-tune the three Czech monolingual BERT-like models, i.e., \textit{Czert-B}, \textit{RobeCzech} and \textit{Czech Electra} model and two multilingual models \textit{mBERT} and \textit{XLM-R}. For English, we use the same two multilingual models and the original \textit{BERT} model. In \textit{cross-lingual} experiments, we test the ability to transfer knowledge between Czech and English using the \textit{zero-shot cross-lingual} classification. We fine-tune the multilingual models only on the dataset in one language (Czech or English) and then evaluate the fine-tuned model on the dataset in the other language.

\par  We always fine-tune\footnote{The composition of data used for training and evaluation depends on the corresponding experiment. In the case of English monolingual experiments for the 10-fold split, we did not use any development data.} on training data and measure the results on the development and testing data parts. We select the model that performs best on the development data and we report the results using average accuracy with the 95\% confidence intervals (we repeat each experiment at least 12 times). We fine-tune all parameters of the model, including the added classification layers. We run the experiments for at most ten epochs with the linear learning rate decay (without learning rate warm-up) with the initial learning rates ranging from 2e-7 to 2e-4. The batch size is set to 32 and the max sequence length of the input is 200 since we classify sentences and the vast majority of them fit into this length. See Appendix \ref{sec:transformer-appendix} for the hyper-parameters details for the reported experiment results.

\subsection{Czech Monolingual Experiments}
\label{sec:czech-monolingual-experiments}
For Czech monolingual experiments, we use two types of training data. The training part (\texttt{cs-train}) of the manually labeled dataset \texttt{Subj-CS} and the entire automatically created dataset \texttt{Subj-CS-L} (marked as \texttt{cs-L-train}). In both cases, we evaluate models on the development (\texttt{cs-dev}) and testing (\texttt{cs-test}) parts of the \texttt{Subj-CS} dataset. We report the results in Table \ref{tab:monolingual-cs}.

\begin{table}[ht!]
\catcode`\-=12
\begin{adjustbox}{width=0.95\linewidth,center}
\begin{tabular}{lccc} \toprule
 \multirow{2}{*}{Model}                       & \multicolumn{1}{c}{Subj-CS (cs-train)}  &   & \multicolumn{1}{c}{Subj-CS-L (cs-L-train)}                                  \\ \cline{2-2} \cline{4-4} 
        & \multicolumn{1}{c}{cs-test}  &       & \multicolumn{1}{c}{cs-test}     \\ \midrule
Czech Electra & 91.85 $\pm$ 0.27\phantom{*}  & & 91.21 $\pm$ 0.08\phantom{*} \\
Czert-B    & 92.85 $\pm$ 0.20\phantom{*}  & & 91.79 $\pm$ 0.07* \\
RobeCzech    & 93.29 $\pm$ 0.19*   & & 91.63 $\pm$ 0.08\phantom{*} \\
mBERT        & 91.23 $\pm$ 0.21\phantom{*}  & & 91.14 $\pm$ 0.11\phantom{*} \\
XLM-R-Large  & \textbf{93.56} $\pm$ \textbf{0.13}\phantom{*}  & & \textbf{91.96} $\pm$ \textbf{0.10\phantom{*}} \\ \bottomrule
\end{tabular}
\end{adjustbox}
\caption{Results for Czech monolingual experiments reported as average accuracy for the testing \texttt{cs-test} data part. The * symbol denotes results containing intersection in confidence interval with the best model.} \label{tab:monolingual-cs}
\end{table}

\par As we expected, the XLM-R-Large model achieves the highest average accuracy of $93.56\%$ for both types of training data. Despite the highest achieved accuracy, there is an intersection in its confidence interval with RobeCzech model for the \texttt{cs-train} data
(the * symbol in Table \ref{tab:monolingual-cs}). Thus, we can conclude that RobeCzech and XLM-R-Large perform very similarly for Czech monolingual experiments. Thanks to the XLM-R-Large size (and its relatively large hardware training requirements), one could prefer the smaller RobeCzech model. The last observation is that all the models achieve better results with the \texttt{cs-train} data part. We expected XLM-R-Large to perform very well because it is the largest model and as shown in \cite{priban-steinberger-2021-multilingual} it usually outperforms smaller monolingual models.

\subsection{English Monolingual Experiments}
\label{sec:english-monolingual-experiments}

\par In our English monolingual experiments, we evaluate the English dataset on our~training (\texttt{en-train}), development (\texttt{en-dev}) and testing (\texttt{en-test}) data split.  Because models from other works \cite{zhao2015self-adasent,CNN-MCFA-subj-2018,khodak-etal-2018-la-byte-mLSTM7,reimers-gurevych-2019-sentence,nandi2021empirical-indove-subj} are evaluated on the 10-fold split, we evaluate the models also on the 10-fold split (\texttt{en-10-fold}) to be able to compare their and ours results.

\begin{table}[ht!]
\catcode`\-=12
\begin{adjustbox}{width=0.99\linewidth,center}
\begin{tabular}{llc} \toprule
Model                  & \multicolumn{1}{c}{en-test} & \multicolumn{1}{c}{en-10-fold} \\ \midrule
BERT             & 96.55 $\pm$ 0.16                 & 96.87 $\pm$ 0.25              \\
mBERT            & 95.87 $\pm$ 0.13                 & 96.03 $\pm$ 0.24              \\
XLM-R-Large            & \textbf{97.28} $\pm$ \textbf{0.07}                 & \textbf{97.34} $\pm$ \textbf{0.21}              \\ \hdashline
\cite{wang2021entailment-few-shot-learner}$\dagger$                    & \multicolumn{1}{l}{97.40 $\pm$ 0.10*}                         &  \multicolumn{1}{c}{-}                           \\

\cite{nandi2021empirical-indove-subj} &   \multicolumn{1}{c}{-}                              & 97.30                        \\

\cite{zhao2015self-adasent}               & \multicolumn{1}{c}{-}                              & \multicolumn{1}{c}{95.50}                        \\
\cite{CNN-MCFA-subj-2018}               &  \multicolumn{1}{c}{-}                              & \multicolumn{1}{c}{94.80}                        \\
\cite{khodak-etal-2018-la-byte-mLSTM7}            &  \multicolumn{1}{c}{-}                              & \multicolumn{1}{c}{94.70} \\

\cite{reimers-gurevych-2019-sentence}        &  \multicolumn{1}{c}{-}                              & \multicolumn{1}{c}{94.52} \\ 
\bottomrule                     
\end{tabular}
\end{adjustbox}
\caption{Results for English monolingual experiments reported as average accuracy for the testing \texttt{en-test} and \texttt{en-10-fold} data parts. The model in paper marked with the $\dagger$ symbol uses the same test size, but the distribution of sentences is different in each split part and they also use the standard deviation instead of the confidence interval.} \label{tab:monolingual-en}
\end{table}

\par As shown in Table \ref{tab:monolingual-en}, the XLM-R-Large performs best among the other two transformer models without any intersection of confidence intervals between the different models. We can also see that the results for \texttt{en-test} and \texttt{en-10-fold} are very similar and their confidence intervals overlap for the same model pairs (but different training data). 
Based on this observation, we assume that the results for \texttt{en-test} and \texttt{en-10-fold} are comparable to each other, thus in the cross-lingual experiments, English is evaluated only on the \texttt{en-test} part. 

\par We compare our results with the current state-of-the-art results (rows below the dashed line in Table \ref{tab:monolingual-en}). Most of the other works use the 10-fold cross-validation and our results also achieve the SotA results and are on par with them. We have to note that our 10-fold splits are not exactly the same as those in the referenced works because the authors do not provide them publicly. Using their distribution, we would probably get slightly different results. Nevertheless, we believe that we can compare our results with the other works to some extent.


\subsection{Cross-lingual Experiments}
We perform three types of cross-lingual experiments: from English to Czech, from Czech to English and joint training and evaluation of both languages. The first two are also known as a zero-shot cross-lingual classification because the model is fine-tuned only on data from one language (source language) and evaluated on data from the second language (target language). The model has never seen the labeled data from the target language.

\par For the experiments from English to Czech (\texttt{EN$\rightarrow$CS}), we fine-tune the multilingual models on English \texttt{en-train} data and we evaluate them on the \texttt{en-dev} and \texttt{cs-test}. We select the model that performs best on the \texttt{en-dev} (i.e., the same best model as for the English monolingual data) and we report results for the \texttt{cs-test} data in Table \ref{tab:crosslingual-en-cs}\footnote{We also include the monolingual results for an easier comparison of the results.}.

\begin{table}[h!]
\begin{adjustbox}{width=1\linewidth,center}
\begin{tabular}{lllcc} \toprule
\multirow{2}{*}{Model}            & \multicolumn{2}{c}{EN $\rightarrow$ CS} & &  \multicolumn{1}{c}{ Monoling. (cs-train)}  \\ \cline{2-3} \cline{5-5} 
        & \multicolumn{1}{c}{en-dev}           & \multicolumn{1}{c}{cs-test} & & \multicolumn{1}{c}{cs-test}       \\ \midrule
mBERT & 95.38 $\pm$ 0.22     & 86.18 $\pm$ 0.33  &   & 91.23 $\pm$ 0.21 \\
XLM-R-Large & 97.60 $\pm$ 0.18     & \textbf{90.75} $\pm$ \textbf{0.32}   &  & 93.56 $\pm$ 0.13 \\ \bottomrule
\end{tabular}
\end{adjustbox}
\caption{Accuracy results for cross-lingual experiments from English to Czech along with the results for models trained on monolingual data.} \label{tab:crosslingual-en-cs}
\end{table}

\par The XLM-R-Large model clearly outperforms the mBERT model by $4.5\%$ but is worse than the same model that was trained on monolingual data roughly by $2.8\%$. In the case of mBERT, the results are much worse ($5\%$ difference) than the model trained only on monolingual data.

For experiments from Czech to English (\texttt{CS$\rightarrow$EN}), we fine-tune the models on \texttt{cs-train} and evaluate on \texttt{cs-dev} and \texttt{en-test}. We select the model that performs best on \texttt{cs-dev}.

\par We also train the model on the \texttt{cs-L-train} data, but in this case, we select the model that performs best on the \texttt{en-dev} data from the target language (English). We use the \texttt{en-dev} for selecting the best model because we found out that if we use \texttt{cs-L-dev}, we get much worse results (up to 20\% worse) for the \texttt{en-test}. We are aware of this simplification of the zero-shot cross-lingual classification task, but otherwise, we would not be able to obtain a model with reasonable results. The results are stated in Table \ref{tab:crosslingual-cs-en}.

\begin{table*}[h!]
\begin{adjustbox}{width=0.82\linewidth,center}
\begin{tabular}{lccccccc} \toprule
 \multirow{2}{*}{Model}             & \multicolumn{2}{c}{CS $\rightarrow$ EN (cs-train)} & & \multicolumn{2}{c}{CS $\rightarrow$ EN (cs-L-train)}  & & \multicolumn{1}{c}{Monolingual (en-train)} \\ \cline{2-3} \cline{5-6} \cline{8-8}
     & cs-dev           & en-test    &      & en-dev           & en-test       &  & en-test         \\ \midrule
mBERT & 92.11 $\pm$ 0.38     & 88.99 $\pm$ 0.94 &      &        85.80 $\pm$ 0.89                   &    85.53 $\pm$ 0.98            & &  95.87 $\pm$ 0.13  \\
XLM-R-Large & 94.40 $\pm$ 0.36     & \textbf{92.86} $\pm$ \textbf{0.44} & &         93.35 $\pm$ 0.22                     &       \textbf{90.98} $\pm$ \textbf{0.26}           &  &  97.28 $\pm$ 0.07  \\ \bottomrule
\end{tabular}
\end{adjustbox}
\caption{Accuracy results for cross-lingual experiments from Czech to English along with the results for models trained on monolingual data.} \label{tab:crosslingual-cs-en}
\end{table*}

\par For both models trained on Czech data (\texttt{cs-train} and \texttt{cs-L-train}), the results are even worse in comparison to the previous experiment from English to Czech. For example, the difference between XLM-R-Large trained on \texttt{cs-train} and XLM-R-Large trained on English \texttt{en-train} data is $4.4\%$, whereas in the case of the previous experiment from English to Czech, it was only $2.8\%$. The results of the models trained on the \texttt{cs-L-train} are significantly worse ($10\%$ for mBERT).

\begin{table*}[ht!]
\begin{adjustbox}{width=0.82\linewidth,center}
\begin{tabular}{lcccccc} \toprule
            & \multicolumn{2}{c}{Joint (cs-train + en-train)} && \multicolumn{1}{c}{Monolingual (cs-train)} &&  Monolingual (en-train) \\ \cline{2-3} \cline{5-5} \cline{7-7}
Model       & cs-test        & en-test       && cs-test            && en-test             \\ \midrule
mBERT       & 91.12 $\pm$ 0.24   & 95.69 $\pm$ 0.22  && 91.23 $\pm$ 0.21 && 95.87 $\pm$ 0.13 \\
XLM-R-Large & \textbf{93.85} $\pm$ \textbf{0.15}   & \textbf{96.95} $\pm$ \textbf{0.12}  && 93.56 $\pm$ 0.13 && 97.28 $\pm$ 0.07 \\ \bottomrule
\end{tabular}
\end{adjustbox}
\caption{Accuracy results for models jointly trained on English and Czech data along with the results for models trained on monolingual data.} \label{tab:crosslingual-joint}
\end{table*}

\par Finally, we fine-tune the models jointly on \texttt{cs-train} and \texttt{en-train}, i.e., on both languages at once. We average the results obtained on \texttt{cs-dev} and \texttt{en-dev} and we select the model that achieves the highest average value. We report the results for the \texttt{cs-test} and \texttt{en-test} in Table \ref{tab:crosslingual-joint}.

\par We can see that the obtained results are almost identical or slightly different compared to the models trained only on monolingual data. Thus, we can conclude that the joint fine-tuning has no beneficial contribution.

\subsection{Discussion}
In this section, we summarize and mention some of our main findings and conclusions from the experiments. Even though that the Czech Electra model is significantly smaller than all the other models, it achieves very competitive results compared to the other models. Thanks to its smaller size, it is much easier and faster to be fine-tuned. 

\par The XLM-R-Large model dominates the results, but it is also several times larger than the other models, see Table \ref{tab:models-size}. Despite the worse results in the cross-lingual experiments, we can state that generally, the XLM-R-Large (and in some cases even mBERT) is relatively capable of transferring knowledge between Czech and English and vice versa, at least for the subjectivity classification task.
\textcolor{black}{The confidence intervals for results obtained in cross-lingual experiments are usually larger than the ones for the monolingual results. Thus, we consider the cross-lingual results less stable.}

\par During the cross-lingual experiments, we select the best model based on development results for the source language. We believe that this is more difficult and challenging than choosing the model according to the results on the target language. We also believe that this setting is much closer to the potential usage of the multilingual models in the industry or to solving practical real-world tasks that are often more complicated. We do not use this approach for models trained on the large data that were obtained automatically because of its poor results.

\par Based on the cross-lingual results, we believe that for knowledge transfer between languages, a smaller but high-quality (manually annotated) dataset is better and more important than a large automatically created dataset to obtain more reliable results for downstream tasks.







\section{Conclusion}
In this work, we introduce the first Czech subjectivity dataset \texttt{Subj-CS} that consists of 10k manually annotated subjective and objective sentences from movie reviews and descriptions.
In addition, we automatically compiled a second much larger dataset of 200k sentences. Both datasets are freely available for research purposes.

\par We describe the process of building and annotating the dataset. The dataset was annotated by two annotators with Cohen’s $\kappa$ inter-annotator agreement equal to $0.83$. In the paper, we provide a summary of the annotation guidelines used by the annotators.

\par We perform a series of monolingual experiments with five pre-trained BERT-like models to obtain the baseline results for the newly created Czech dataset and we are able to achieve $93.5\%6$ of accuracy with the XLM-R-Large model. We also perform monolingual experiments for the existing English subjectivity dataset with three models obtaining $97.28\%$ of accuracy, which is on par with the current state-of-the-art results for this dataset. Finally, we conduct zero-shot cross-lingual subjectivity classification to verify the usability of our dataset as the cross-lingual benchmark for pre-trained multilingual models that allow transfer learning. 

\par Our experiments confirm that we provide a dataset of relatively high quality and it can be used as an evaluation benchmark to test the ability of pre-trained models to transfer knowledge between Czech and English.

\par In future work, we want to focus on using the dataset to improve sentiment analysis (polarity detection) in Czech and English. Furthermore, we would like to include sentences labeled as \textit{unsure} in the dataset, along with a detailed error analysis of the fine-tuned models.





\section{Acknowledgments}
This work has been partly supported by ERDF ”Research and Development of Intelligent Components of Advanced Technologies for the Pilsen Metropolitan Area (InteCom)” (no.: CZ.02.1.01/0.0/0.0/17 048/0007267); and by Grant No. SGS-2022-016 Advanced methods of data processing and analysis. Computational resources were supplied by the project "e-Infrastruktura CZ" (e-INFRA CZ LM2018140 ) supported by the Ministry of Education, Youth and Sports of the Czech Republic.


\section{Bibliographical References}\label{reference}

\bibliographystyle{lrec2022-bib}
\bibliography{lrec2022-example}


\appendix

\section{Appendix}
\subsection{Hyper-parameters}
\label{sec:transformer-appendix}
During fine-tuning, we tried a variety of hyper-parameters, we use the Adam \cite{Kingma-adam} optimizer with default parameters ($\beta_1 = 0.9, \beta_2 = 0.999$) and the cross-entropy loss function. We randomly shuffle training data before each epoch. We run the experiments for at most ten epochs with the linear learning rate decay (without learning rate warm-up) with the initial learning rates ranging from 2e-7 to 2e-4. The 2e-4 learning rate was used only for the Czech Electra model, when used with other models, the models started to diverge. The batch size is always set to 32 and the max length of the input sequence is 200. In Tables \ref{tab:hyper-monolingual-cs}, \ref{tab:hyper-monolingual-en}, \ref{tab:hyper-crosslingual-en-cs}, \ref{tab:hyper-crosslingual-cs-en} and \ref{tab:hyper-crosslingual-joint} we report results with the used initial learning rate and a number of epochs in parentheses. The first number in brackets is the initial learning rate and the second is the number of epochs for fine-tuning.

\begin{table}[H]
\catcode`\-=12
\begin{adjustbox}{width=1\linewidth,center}
\begin{tabular}{lccc} \toprule
 \multirow{2}{*}{Model}                       & \multicolumn{1}{c}{Subj-CS (cs-train)}  &   & \multicolumn{1}{c}{Subj-CS-L (cs-L-train)}                                  \\ \cline{2-2} \cline{4-4} 
        & \multicolumn{1}{c}{cs-test}  &       & \multicolumn{1}{c}{cs-test}     \\ \midrule
Czech Electra & 91.85 $\pm$ 0.27\phantom{*} \footnotesize{(2e-4 / 4)}  & & 91.21 $\pm$ 0.08\phantom{*} \footnotesize{(2e-5 / 7)} \\
Czert-B    & 92.85 $\pm$ 0.20\phantom{*} \footnotesize{(2e-5 / 3)}  & & 91.79 $\pm$ 0.07* \footnotesize{(2e-6 / 7)} \\
RobeCzech    & 93.29 $\pm$ 0.19* \footnotesize{(2e-5 / 7)}   & & 91.63 $\pm$ 0.08\phantom{*} \footnotesize{(2e-6 / 2)} \\
mBERT        & 91.23 $\pm$ 0.21\phantom{*} \footnotesize{(2e-5 / 3)}  & & 91.14 $\pm$ 0.11\phantom{*} \footnotesize{(2e-6 / 5)} \\
XLM-R-Large  & \textbf{93.56} $\pm$ \textbf{0.13}\phantom{*} \footnotesize{(2e-5 / 4)}  & & \textbf{91.96} $\pm$ \textbf{0.10\phantom{*}} \footnotesize{(2e-6 / 9)} \\ \bottomrule
\end{tabular}
\end{adjustbox}
\caption{Results with model hyper-parameters for Czech monolingual experiments reported as average accuracy for the testing \texttt{cs-test} data part. The * symbol denotes results containing intersection in confidence interval with the best model.} \label{tab:hyper-monolingual-cs}
\end{table}

\begin{table*}[ht!]
\begin{adjustbox}{width=0.95\linewidth,center}
\begin{tabular}{lccccccc} \toprule
 \multirow{2}{*}{Model}             & \multicolumn{2}{c}{CS $\rightarrow$ EN (cs-train)} & & \multicolumn{2}{c}{CS $\rightarrow$ EN (cs-L-train)}  & & \multicolumn{1}{c}{Monolingual (en-train)} \\ \cline{2-3} \cline{5-6} \cline{8-8}
     & cs-dev           & en-test    &      & en-dev           & en-test       &  & en-test         \\ \midrule
mBERT & 92.11 $\pm$ 0.38     & 88.99 $\pm$ 0.94 \footnotesize{(2e-5 / 3)} &      &        85.80 $\pm$ 0.89                   &    85.53 $\pm$ 0.98  \footnotesize{(2e-6 / 1)}           & &  95.87 $\pm$ 0.13  \footnotesize{(2e-5 / 10)}   \\
XLM-R-Large & 94.40 $\pm$ 0.36     & \textbf{92.86} $\pm$ \textbf{0.44} \footnotesize{(2e-5 / 4)} & &         93.35 $\pm$ 0.22                     &       \textbf{90.98} $\pm$ \textbf{0.26}   \footnotesize{(2e-7 / 1)}           &  &  97.28 $\pm$ 0.07 \footnotesize{(2e-6 / 10)} \\ \bottomrule
\end{tabular}
\end{adjustbox}
\caption{Accuracy results with model hyper-parameters for cross-lingual experiments from Czech to English along with the results for models trained on monolingual data.} \label{tab:hyper-crosslingual-cs-en}
\end{table*}

\begin{table*}[t]
\begin{adjustbox}{width=0.90\linewidth,center}
\begin{tabular}{lclcccc} \toprule
            & \multicolumn{2}{c}{Joint (cs-train + en-train)} && \multicolumn{1}{c}{Monolingual (cs-train)} &&  Monolingual (en-train) \\ \cline{2-3} \cline{5-5} \cline{7-7}
Model       & cs-test        & \multicolumn{1}{c}{en-test}       && cs-test            && en-test             \\ \midrule
mBERT       & 91.12 $\pm$ 0.24   & 95.69 $\pm$ 0.22 \footnotesize{(2e-5 / 3)}  && 91.23 $\pm$ 0.21 \footnotesize{(2e-5 / 3)} && 95.87 $\pm$ 0.13 \footnotesize{(2e-5 / 10)} \\
XLM-R-Large & \textbf{93.85} $\pm$ \textbf{0.15}   & \textbf{96.95} $\pm$ \textbf{0.12} \footnotesize{(2e-6 / 10)}  && 93.56 $\pm$ 0.13 \footnotesize{(2e-5 / 4)} && 97.28 $\pm$ 0.07 \footnotesize{(2e-6 / 10)} \\ \bottomrule
\end{tabular}
\end{adjustbox}
\caption{Accuracy results with hyper-parameters for models jointly trained on English and Czech data along with the results for models trained on monolingual data.} \label{tab:hyper-crosslingual-joint}
\end{table*}

\begin{table}[H]
\catcode`\-=12
\begin{adjustbox}{width=1\linewidth,center}
\begin{tabular}{llc} \toprule
Model                  & \multicolumn{1}{c}{en-test} & \multicolumn{1}{c}{en-10-fold} \\ \midrule
BERT             & 96.55 $\pm$ 0.16 \footnotesize{(2e-5 / 3)}                & 96.87 $\pm$ 0.25 \footnotesize{(2e-5 / 9)}             \\
mBERT            & 95.87 $\pm$ 0.13 \footnotesize{(2e-5 / 10)}                & 96.03 $\pm$ 0.24 \footnotesize{(2e-5 / 5)}             \\
XLM-R-Large            & \textbf{97.28} $\pm$ \textbf{0.07} \footnotesize{(2e-6 / 10)}                 & \textbf{97.34} $\pm$ \textbf{0.21} \footnotesize{(2e-5 / 4)}              \\ \bottomrule                     
\end{tabular}
\end{adjustbox}
\caption{Results with model hyper-parameters for English monolingual experiments reported as average accuracy for the testing \texttt{en-test} and \texttt{en-10-fold} data parts.} \label{tab:hyper-monolingual-en}
\end{table}

\begin{table}[H]
\begin{adjustbox}{width=1\linewidth,center}
\begin{tabular}{lllcc} \toprule
\multirow{2}{*}{Model}            & \multicolumn{2}{c}{EN $\rightarrow$ CS} & &  \multicolumn{1}{c}{ Monoling. (cs-train)}  \\ \cline{2-3} \cline{5-5} 
        & \multicolumn{1}{c}{en-dev}           & \multicolumn{1}{c}{cs-test} & & \multicolumn{1}{c}{cs-test}       \\ \midrule
mBERT & 95.38 $\pm$ 0.22     & 86.18 $\pm$ 0.33 \footnotesize{(2e-5 / 10)}   &   & 91.23 $\pm$ 0.21 \footnotesize{(2e-5 / 3)} \\
XLM-R-Large & 97.60 $\pm$ 0.18     & \textbf{90.75} $\pm$ \textbf{0.32} \footnotesize{(2e-6 / 10)}  &  & 93.56 $\pm$ 0.13 \footnotesize{(2e-5 / 4)} \\ \bottomrule
\end{tabular}
\end{adjustbox}
\caption{Accuracy results with model hyper-parameters for cross-lingual experiments from English to Czech along with the results for models trained on monolingual data.} \label{tab:hyper-crosslingual-en-cs}
\end{table}


\end{document}